\documentclass[preprint,12pt]{elsarticle}

\usepackage{graphicx}
\usepackage{amssymb}
\usepackage{amsmath}
\usepackage[ruled,linesnumbered]{algorithm2e}
\usepackage{hyperref}
\usepackage{lineno}

\journal{Swarm and Evolutionary Computation}

\begin{document}

\begin{frontmatter}


\title{Solving Sudoku with Ant Colony Optimisation}



\author{Huw Lloyd and Martyn Amos}

\address{Centre for Advanced Computational Science, Manchester Metropolitan University, Manchester M1 5GD, United Kingdom.}

\begin{abstract}
In this paper we present a new Ant Colony Optimisation-based algorithm for Sudoku, which out-performs existing methods on large instances. Our method includes a novel anti-stagnation operator, which we call Best Value Evaporation.
\end{abstract}


\end{frontmatter}


\section{Introduction}
\label{sec:intro}

Sudoku is a well-known logic-based puzzle game that was first published in 1979 under the name of ``Number Place''. It was popularised in Japan in 1984 by the puzzle company Nikoli, and later named ``Sudoku", which roughly translates to  ``single digits''. The puzzle gained attention in the West in 2004, after {\it The Times} published its first Sudoku grid (at the instigation of Hong Kong-based judge Wayne Gould, who first encountered the puzzle in 1997, and developed a computer program to automatically generate instances). Sudoku is now a global phenomenon, and many newspapers now carry it alongside their existing crosswords (see \cite{delahaye2006science} for a general history of the puzzle).

\begin{figure}[!ht]
\centering
\includegraphics[width=2in]{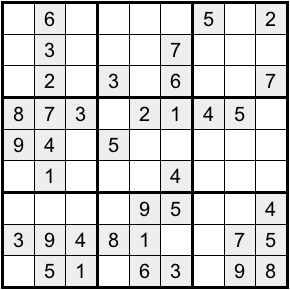}
\hfil
\includegraphics[width=2in]{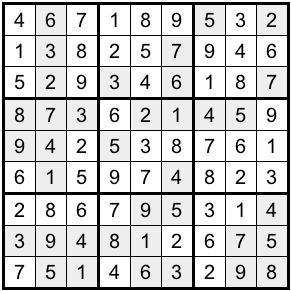}
\caption{The structure of a Sudoku puzzle instance (left), and its solution (right).}
\label{fig:grid}
\end{figure}

The simplest variant of Sudoku uses a 9$\times$9 grid of cells divided into nine 3$\times$3 subgrids (Figure ~\ref{fig:grid} (left)). The aim of the puzzle is to fill the grid with digits such that each row, each column, and each 3$\times$3 subgrid contains all of the digits 1-9 (Figure ~\ref{fig:grid} (right)). An instance of Sudoku provides, at the outset, a partially-completed grid, but the difficulty of any grid derives more from the range of techniques required to solve it than the number of cell values that are provided for the player.

Sudoku is an NP-complete problem \cite{gareyjohnson79}, as first shown in \cite{yato2003complexity} (via a reduction from the Latin Square Completion problem \cite{colbourn1984complexity}). As such, the problem offers itself as a useful benchmark challenge, and a number of different types of algorithm have been proposed for its solution.

The rest of the paper is structured as follows: in Section ~\ref{sec:related}  we briefly review closely-related recent work on the application of various algorithms to Sudoku. This motivates the description, in Section ~\ref{sec:algorithm} of our own method, based on Ant Colony Optimisation (ACO), which introduces a novel operator which we call {\it Best Value Evaporation}. In Section ~\ref{sec:results} we present the results of experimental investigations, which confirm that our algorithm out-performs existing methods, and we conclude in Section ~\ref{sec:conclusions} with a discussion of our findings, and discuss possible future work in this area.

\section{Related work}
\label{sec:related}

The {\it Exact Cover Problem} \cite{karp1972reducibility} is a type of constraint satisfaction problem which may be phrased as follows: given a binary matrix, find a {\it subset} of rows in which each column sums to 1 (that is, find a set of rows in which each column contains only a single 1).
In \cite{knuth2000dancing}, Knuth describes the ``dancing links'' implementation of his Algorithm X (called DLX), a ``brute force" backtracking algorithm for Exact Cover. As any Sudoku puzzle may be transformed into an instance of Exact Cover \cite{hunt2007difficulty}, DLX naturally offers an effective solution method for Sudoku \cite{fletcher2007taking}.

In \cite{norvigsolving}, Peter Norvig presents an alternative approach, based on {\it constraint propagation} followed by a search process (we discuss this in more detail shortly). Other notable approaches to solving Sudoku include formal logic \cite{weber2005sat}, an artificial bee colony algorithm \cite{pacurib2009solving}, constraint programming  \cite{crawford2008using,lewis2007metaheuristics}, evolutionary algorithms \cite{deng2013novel,mantere2007solving,segura2016importance,wang2015evolutionary}, particle swarm optimisation \cite{hereford2008integer,moraglio2007geometric}, simulated annealing \cite{karimi2010sudoku}, tabu search \cite{soto2013hybrid}, and entropy minimization \cite{gunther2012entropy}.

In this paper, we focus on the application of ACO to the solution of Sudoku. ACO is a population-based search method inspired by the foraging behaviour of ants \cite{dorigo1999ant,dorigo1996ant}, and it has been successfully applied to a wide range of computational problems (see \cite{dorigo2011ant,lopez2016ant} for overviews of both the algorithm and its applications).

In \cite{mantere2013improved}, Mantere presents a hybrid ACO/genetic algorithm approach to Sudoku, which combines global (evolutionary) search with greedy local (ACO-based) search. Schiff \cite{schiff2015ant} and Sabuncu \cite{sabuncu2015work} also present relatively recent work on applying ACO to Sudoku, but, in both cases, the performance of the algorithm is relatively poor. 

For the purposes of comparison, in this paper we focus mainly on the work of Musliu, {\it et al.} \cite{musliu2017hybrid}, who present an iterated local search algorithm with constraint programming which represents the state-of-the-art in stochastic search algorithms for the Sudoku problem, plus the algorithms of Knuth and Norvig \cite{knuth2000dancing,norvigsolving}.

\section{Our algorithm}
\label{sec:algorithm}

In \cite{norvigsolving}, Norvig describes a two-component approach to solving Sudoku, using a combination of {\it constraint propagation} (CP) and {\it search}. CP ensures that the ``rules'' of Sudoku are observed, and repeatedly prunes the {\it value set} of each cell (that is, the set of possible values that cells might take). Importantly, by using CP during search, we effectively ``parallelise'' the process, by eliminating large numbers of possible cell values every time we fix a cell's value (because selecting a specific value for a cell immediately rules out that value's presence in a large number of other cells). In \cite{norvigsolving}, Norvig combines CP with a recursive depth-first search which, at each iteration, selects the cell with the smallest value set (which essentially maximises the probability of ``guessing correctly"), and then chooses the first value (in numeric order) for that cell; this is the {\em Minimum Remaining Values Heuristic}.

Here, we present a variant of constraint propagation inspired by Norvig's method, and use ACO (rather than depth-first search) to search the space of solutions. We now describe our CP method in more detail. For clarity, this is written in terms of the $9\times 9$ Sudoku puzzle, but the method generalises trivially to larger sizes (e.g. $16\times 15$, $25\times 25$).

\newpage
\subsection{Constraint propagation}
\label{sec:constraints}

A Sudoku problem is made up of a grid of {\it cells} (or squares), arranged into 3$\times$3 subgrids known as {\it boxes}. A {\it unit} is a row, column or box, each containing exactly nine cells. A problem is solved when {\it each unit} (that is, every row, column and box) contains a permutation of the digits 1$\dots$9 \cite{norvigsolving}

\begin{figure}[!ht]
\centering
{\includegraphics[width=5.5in]{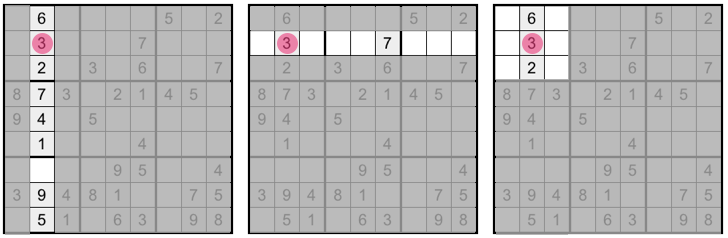}}
\caption{Units and peers for a specific highlighted cell.}
\label{fig:unitspeers}
\end{figure}

Any given cell has exactly three units and 20 {\it peers}; the units are the row, column and box in which the cell resides, and the set of peers is made up of the other cells in those units (that is, 2$\times$8=16 neighbours in the relevant row and column, plus 4 other cells occupying the same box; see Figure ~\ref{fig:unitspeers}). Throughout the CP process, each cell maintains its value set - a list of {\it possible values} it might take; every cell starts with the same value set, $[1\dots 9]$. Once a set has been reduced to a single value, we call that value {\it fixed} for that cell. Our CP algorithm implements two basic rules, which are applied to a cell's peers when it has its value fixed:
\begin{enumerate}
\item Eliminate from a cell's value set all values that are {\it fixed} in any of the cell's {\it peers}.
\item If any values in a cell's value set are in the only possible place {\it in any of the cell's units}, then fix that value.
\end{enumerate}
Note that since this can lead to other cells having their values fixed, the procedure is recursive, and terminates when no further changes are possible.

\begin{figure}[!ht]
\centering
{\includegraphics[width=2in]{grida}}
\hfil
{\includegraphics[width=3in]{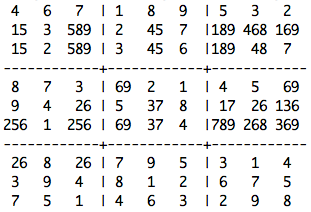}}
\caption{Instance from Figure ~\ref{fig:grid} (left), and (right) cell value sets after initial pass of constraint propagation algorithm.}
\label{fig:instanceCP}
\end{figure}

In Figure ~\ref{fig:instanceCP} we show the instance from Figure ~\ref{fig:grid} after the {\it initial} pass of our CP algorithm (which occurs when the board is set up, before any search is performed). For easy cases, the application of the CP algorithm is often sufficient to solve the board, and no further search is required (see Section ~\ref{sec:results} for a discussion). However, in most cases, some search will be required, and we now describe our ACO-based method for this.

\subsection{Our ACO algorithm}

Our algorithm is based on Ant Colony System (ACS), which is a variant of ACO introduced in \cite{dorigo1997ant}. We first give an informal description of the algorithm, and then formally specify its various components.

At each iteration, each ant starts with a ``fresh'' copy of the puzzle, and the aim of each ant is to fix as many cell values as possible. Each ant starts on a different, randomly-selected cell, and then iterates over all cells on the board. Whenever it leaves a cell that does not have a fixed value (that is, a cell with a number of possible values), an ant must make a decision on which element of that cell's value set to choose (thus setting the cell to that value). Importantly, as soon as an ant sets the value of a cell, the constraints that it introduces are propagated across the board. 

Decisions on which value to choose are based on relative pheromone levels, which are assigned to each possible value. These are stored in a {\it pheromone matrix}, which keeps track of a single pheromone amount for each possible value in each cell. This is, for an order-3 ($9\times 9$) Sudoku puzzle, a matrix of $81 \times 9$ values, with each cell corresponding to the pheromone level for each possible value ($1\ldots 9$) in a cell (indexed $1\ldots 81$). Depending on the ``greediness'' of the selection, either the value with the highest pheromone value is chosen, or a weighted (roulette) selection is made. 
\newpage

\begin{algorithm}[H]
\SetAlgoLined
 read in puzzle \;
 \For{all cells with fixed values}{
 propagate constraints (according to Section ~\ref{sec:constraints}) \;
 }
 initialize global pheromone matrix\;
 \While{puzzle is not solved}{
  give each ant a local copy of puzzle \;
  assign each ant to a different cell \; 
  \For{number of cells}{
  	\For{each ant}{
    	\If{current cell value not set}
    	{
    		choose value from current cell's value set \;
        	set cell value \;
        	propagate constraints \;
        	update local pheromone \;
    		
    	}
        move to next cell \;
    }
  }
  find best ant \;
  do global pheromone update \;
  do best value evaporation
 }
 \caption{Our ACO algorithm for Sudoku}
 \label{alg:algorithm}
\end{algorithm}

After the cell's value is set, the standard ACS {\it local} pheromone operator is applied, which reduces the probability of that value being selected by the following ant (thus preventing early convergence). 

Once all ants have covered every square of the board, we then perform the {\it global} pheromone update, which rewards only the best solution found (in line with ACS principles). We characterise the ``best'' solution, at each iteration, as the sequence of value selections that lead to the greatest number of cells having their values fixed (that is, the best solution is effectively found by the ant that ``guesses'' correctly the highest number of times). At this point, we introduce a variation to the standard ACS algorithm, which we call {\it best value evaporation} (BVE). In standard ACS, the global pheromone operator increases the pheromone concentrations of all components of the global best solution with an amount of pheromone that is directly proportional to the {\it absolute quality} of that solution.  However, this can gradually lead to {\it stagnation}, where all ants end up selecting the same route (citation needed). Instead, the amount of pheromone that is added globally (which we call the {\it best value} is measured in terms of the {\it proportionate quality} of the best solution; that is a new ``best ant'' is an ant that proportionately adds more pheromone than the {\it current} best ant. Importantly, the best value itself is subject to evaporation over time, which prevents ``lock in''; taken together, these two components of BVE prevent premature stagnation (which is confirmed by our later experimental observations).

We give a pseudo-code description of our approach in Algorithm ~\ref{alg:algorithm}, components of which we now formally specify.

\noindent
{\bf Line 5:} For a Sudoku puzzle of dimension $d$ we define a two-dimensional global pheromone matrix, $\tau$, in which each element is denoted as $\tau^{k}_{i}$, where $i$ is the cell index 
$(1 \leq i \leq d^2)$ and $k$ is a possible value for the cell ($k \in [1,d]$). $\tau_i^k$represents the pheromone level associated with value $k$ in cell $i$. Each element of the matrix is initialised to some fixed value, $\tau_0$ (we use a value of $1/c$, where $c = d^2$ is the total number of cells on the board).

\noindent
{\bf Line 12:} Where an ant has a choice of a number of values in an ``open'' cell (i.e., one which does not yet have its value fixed), then we define the {\it value set}, $v_{i}$ of cell $i$ as the set of all available values for that cell, from which we have to choose one. We have a choice of two methods to use when making a selection; we might make a {\it greedy} selection, in which case the member of $v$ with the highest pheromone concentration is selected:

\begin{equation}
g(v)=\operatorname*{argmax}_{n \in v_{i}} \{\tau^{n}_{i}\}
\label{eq:greedy}
\end{equation}

\noindent
or we might make a {\it weighted} (i.e., ``roulette wheel'') selection, in which case the ``weighted probability", $wp$, of value $s \in v$ in cell $i$ being selected is denoted as

\begin{equation}
wp(s) = \frac{\tau^{s}_{i}}{\sum\limits_{n \in v_{i}}^{} \tau^{n}_{i}}
\label{eq:roulette}
\end{equation}

The relative probabilities of each type of selection are determined by the {\it greediness} parameter, $q_0$ $(0 \leq q_0 \leq 1)$, where $0 \leq q \leq 1$ is a uniform random number. A value selection, $s$, is therefore made, as follows:

\begin{equation}
s=
\begin{cases}
    g										& \text{if } q > q_0\\
    Equation ~\ref{eq:roulette}            & \text{otherwise}
\end{cases}
\label{eq:conditional}
\end{equation}

\noindent
{\bf Line 15: } The local pheromone update is handled as follows; every time an ant selects a value, $s$, at cell $i$, its pheromone value in the matrix is updated as follows:

\begin{equation}
\tau^{s}_{i} \leftarrow (1-\xi)\tau^{s}_{i} + \xi \tau_0 
\label{eq:local}
\end{equation}
with $\xi = 0.1$ (the standard setting for ACS).

\noindent
{\bf Line 20: } In order to perform the global pheromone update, we must first find the best-performing ant. At each iteration, each of the $m$ ants keeps track of the number of cells, $f$, that it has managed to set to a specific value.  We then calculate the amount of pheromone to add, $\Delta\tau$, as follows:

\begin{equation}
\Delta\tau \leftarrow \frac{c}{c - \operatorname*{max}\{ f_n\}}
\label{eq:bestant}
\end{equation}

\noindent
{\bf Line 21: } If the value of $\Delta\tau$ exceeds the current ``best pheromone to add'' value, $\Delta\tau_{\rm best}$, then we set $\Delta\tau_{\rm best} \leftarrow \Delta\tau$. We then update all pheromone values corresponding to values in the current best solution, where $\rho$ is the standard evaporation parameter $(0 \leq \rho \leq 1)$:

\begin{equation}
\tau^{s}_{i} \leftarrow (1-\rho)\tau^{s}_{i}  + \rho \Delta\tau_{\rm best}.
\label{eq:global}
\end{equation}
Note that in ACS, there is no global evaporation of pheromone; the global pheromone update (equation~\ref{eq:global}) is only applied to pheromone values corresponding to fixed values in the best solution.

\noindent
{\bf Line 22: } In order to prevent ``lock in'', we then evaporate the current best pheromone value, where $0 \leq f_{BVE} \leq 1$:

\begin{equation}
\Delta\tau_{\rm best} \leftarrow \Delta\tau_{\rm best} \times (1-f_{\rm BVE})
\label{eq:bve}
\end{equation}

\section{Experimental results}

\label{sec:results}

Our ant colony algorithm (ACS) was evaluated by comparing it with (1) iterated local search 
code from Musliu {\it et
al.} (ILS) \cite{musliu2017hybrid}, (2) a C++ implementation of the
Dancing Links algorithm (DLX) \cite{dlx-cpp}, and (3) our own implementation of
backtracking search, using the minimum remaining values heuristic,
which uses the same problem representation and constraint propagation
code as the ant colony algorithm (BS). The code presented in
\cite{musliu2017hybrid} was itself compared against a number of other
stochastic algorithms, and was shown to be the best performing. We include
the Dancing Links and backtracking algorithms for comparison with
deterministic, exhaustive search. Furthermore, including a
backtracking search which uses the same underlying constraint
propagation code allows us to evaluate the effectiveness of the ant
colony algorithm in searching the problem space, independent of the
details of the underlying implementation.

\subsection{Experimental environment}


All of the codes were compiled using the same compiler and
optimisation setting (g++ v4.9.0 with -O3). 
Experiments were run on a machine with an Intel Core-i7 3770 processor with a
clock speed of 3.4GHz, running Debian Linux. The parameter settings
for the iterated local search solver (ILS) were taken from the
recommendations given in \cite{musliu2017hybrid}. For the ant colony code (ACS),
we used the following settings: $\rho=0.9$, $q_0=0.9$, $f_{BVE} =
0.005$, $m=10$. Our code, and all the instance files used for the experiments,
may be downloaded from \url{https://github.com/huwlloyd-mmu/sudoku_acs}.

\subsection{Logic-solvable $9\times 9$ instances}
\label{sec:exp_logic}

\begin{figure}
\begin{center}
\includegraphics[width=\textwidth]{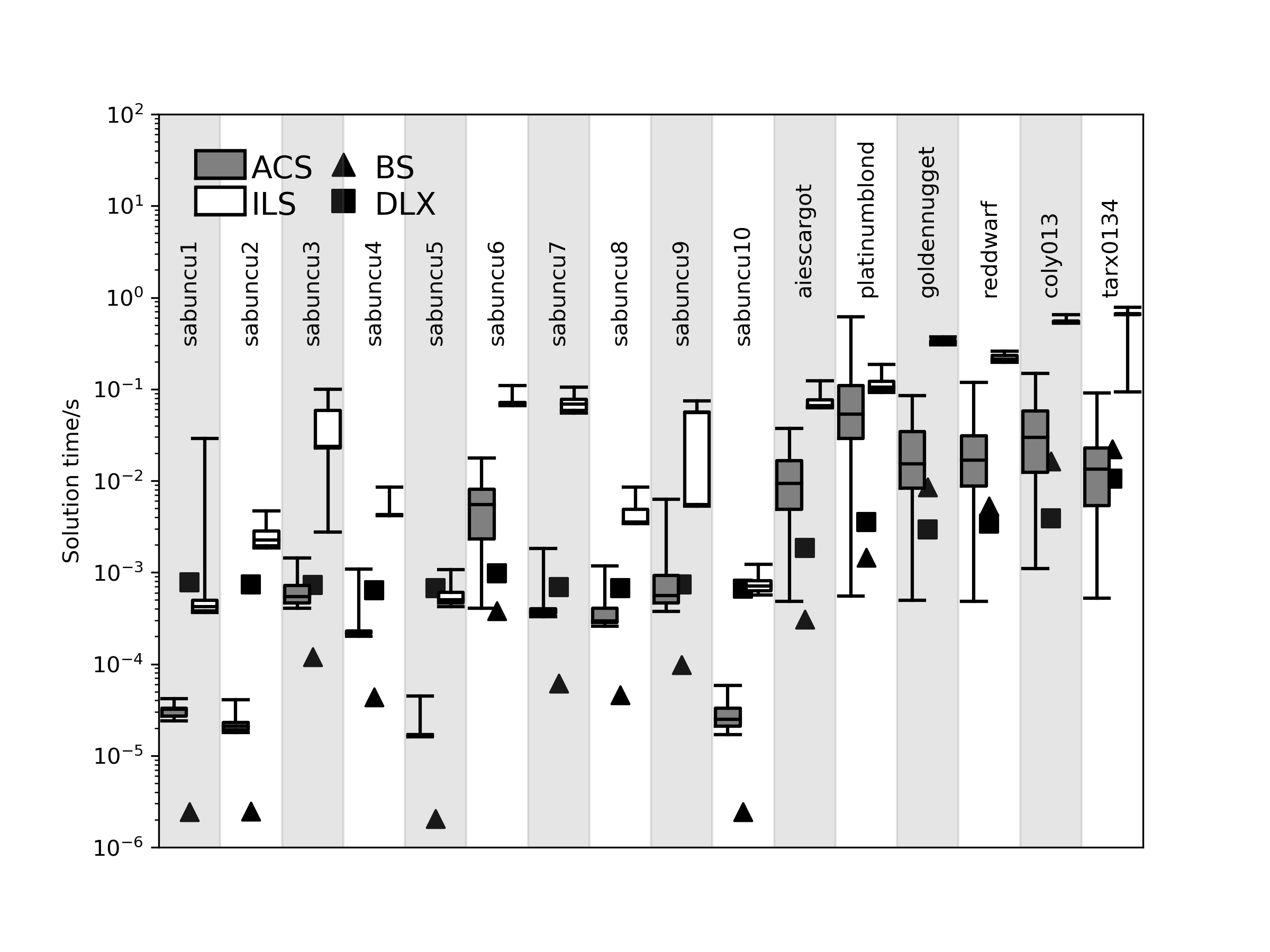}
\caption{Solution times over 100 runs for ACS, ILS, BS and DLX for the sixteen named logic-solvable $9\times 9$ instances. In all cases
ACS is plotted on the left side of the band corresponding to an instance, and ILS (the previous best stochastic algorithm) on the right.}
\label{fig:boxplot}
\end{center}
\end{figure}

We first selected instances based on known difficulty, or on previous use in
the literature. We selected the ten instances used in
\cite{sabuncu2015work} (labelled here {\em sabuncu1} to {\em
  sabuncu10}), five named instances identified by
\cite{ercsey-ravasz2012} as the most difficult ({\em Platinum Blond,
  Golden Nugget, Red Dwarf, coly013, tarx0134}), and one instance ({\em
  AI Escargot}) \cite{aiescargot}, commonly regarded as an extremely
difficult puzzle. These instances are all {\em logic solvable}; in other
words, they each have a unique solution which can be deduced from the
given numbers. We ran the ACS, Iterated Local Search (ILS),
Dancing Links (DLX) and backtracking search (BS) algorithms 100 times
on each instance, with a timeout of 5 seconds.  The puzzles were
successfully solved in all cases by all four algorithms; there were no
time-outs. Figure \ref{fig:boxplot} shows the timing results for the
four algorithms: For ACS and ILS, we give boxplots for the
distribution of times (the boxes represent the quartiles, the whiskers
the minimum and maximum, and the central line the median). For DLX and
BS, the average runtime is given, since these algorithms are
deterministic, and we should not therefore show a distribution of run times.

The ten puzzles from \cite{sabuncu2015work} ({\em
  sabuncu1--sabuncu10}) are generally solved in less time by all the
algorithms than the six harder puzzles.  In four cases
({\em sabuncu1, sabunuc2, sabuncu5} and {\em sabuncu10}) the puzzle is
solved by a single application of our constraint propagation
procedure, so that no searching is required for either the ACS or BS
algorithms. The difference in runtimes between the two algorithms for
these instances may be explained by the difference in {\it set-up}
times; in the case of ACS, the overhead of creating the ant colony and
initializing the pheromone matrix is clearly significant. On these
four ``trivial" instances, the BS algorithm is the fastest of all
(running in times of order a microsecond). DLX requires at least of
order a millisecond to solve all the puzzles; in all but the most
difficult cases, the time is most likely dominated by the calculations
to convert the instance to and from an instance of the exact cover
problem.

ACS is competitive with DLX for most of the instances, and in most
cases the median time is at least an order of magnitude less than the
median time for ILS, although the variation in times seems to be
greater for ACS. We note that the times recorded by
\cite{sabuncu2015work} (typically 1 to 3 seconds) are several
orders of magnitude slower than our times using ACS for the same
instances (of the order of milliseconds, or less); this is more than
can be accounted for by differences in hardware or efficiency of
implementation.

\subsection{General instances}
\label{sec:exp_gen}

Following \cite{lewis2007metaheuristics} and \cite{musliu2017hybrid},
we generated random instances for the $9\times 9$, $16\times 16$ and
$25\times 25$ Sudoku problem. These were generated by running the ACS
code with an initially blank grid, to produce a set of Sudoku
solutions.  These are then converted into problem instances by
randomly blanking a number of the cells. The instances generated in
this way are not guaranteed to have a unique solution. For each of the
sizes $9 \times 9$, $16 \times 16$ and $25 \times 25$, we generated
100 instances for fixed cell fractions in steps of $0.05$ from $0$ to
$0.95$, giving a total of $6000$ individual instances. We ran the ACS,
ILS, DLX and BS codes once on each instance, with timeouts set to 5
seconds for the $9\times 9$ instances, 20 seconds for $16 \times 16$
and 120 seconds for $25 \times 25$.  These timeouts are shorter than
those used by \cite{musliu2017hybrid}; however we ran our experiments on a
faster processor, and with compiler optimisations enabled. Taken
together, these two differences should amount to a factor of
approximately 3 in time.  We designed the experiment so that each
instance is used for one run; this is preferable to carrying out
multiple runs on each of a smaller number of instances
\cite{Birattari2004}.

\begin{figure}
\begin{center}
\includegraphics[width=\textwidth]{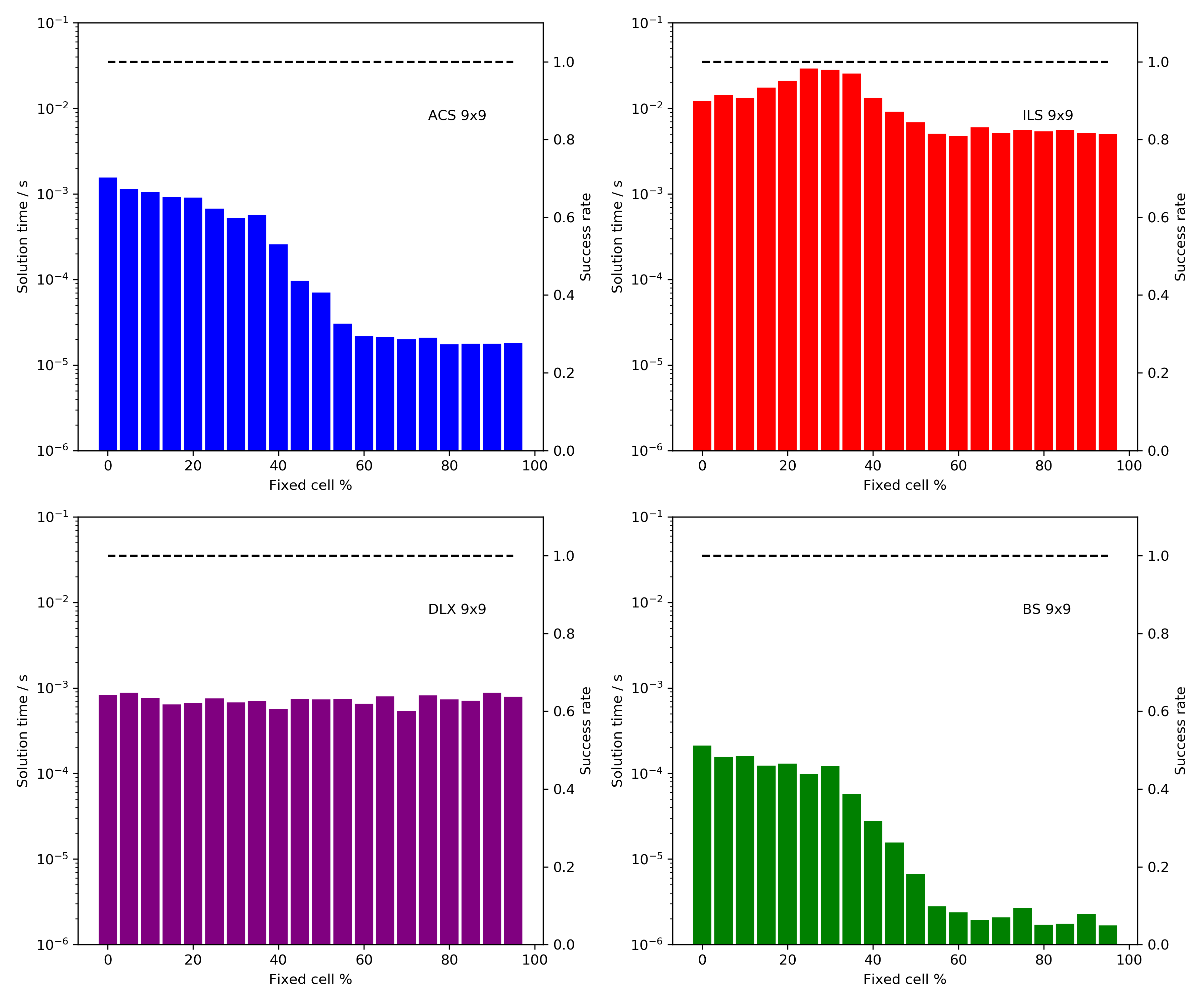}
\caption{Plots of solution time (bars) and success rate (dashed line) against fixed cell percentage for ACS, ILS, BS and DLX for the $9\times 9$ general instances.}
\label{fig:gen9x9}
\end{center}
\end{figure}

\begin{figure}
\begin{center}
\includegraphics[width=\textwidth]{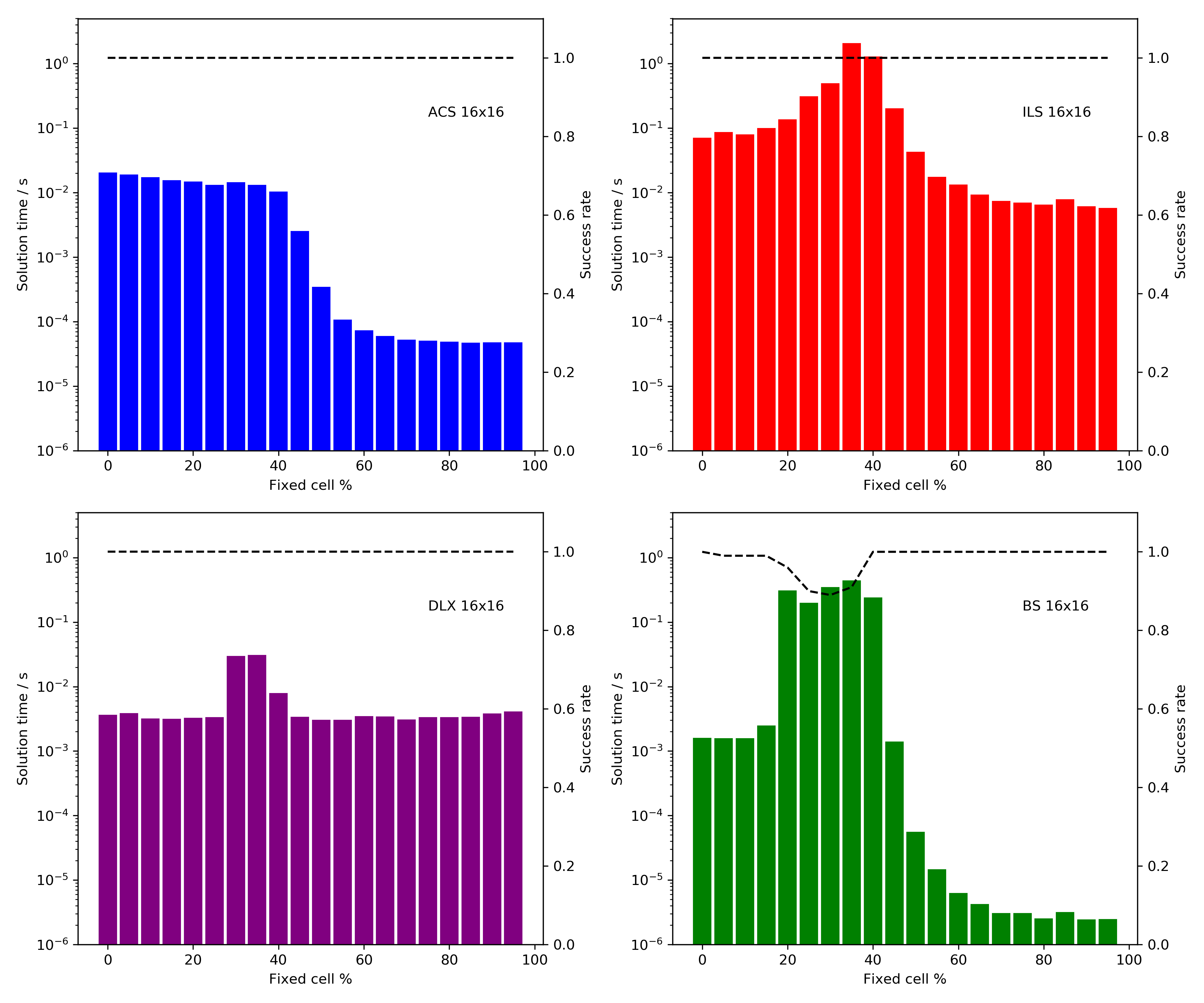}
\caption{Plots of solution time (bars) and success rate (dashed line) against fixed cell percentage for ACS, ILS, BS and DLX for the $16\times 16$ general instances.}
\label{fig:gen16x16}
\end{center}
\end{figure}

\begin{figure}
\begin{center}
\includegraphics[width=\textwidth]{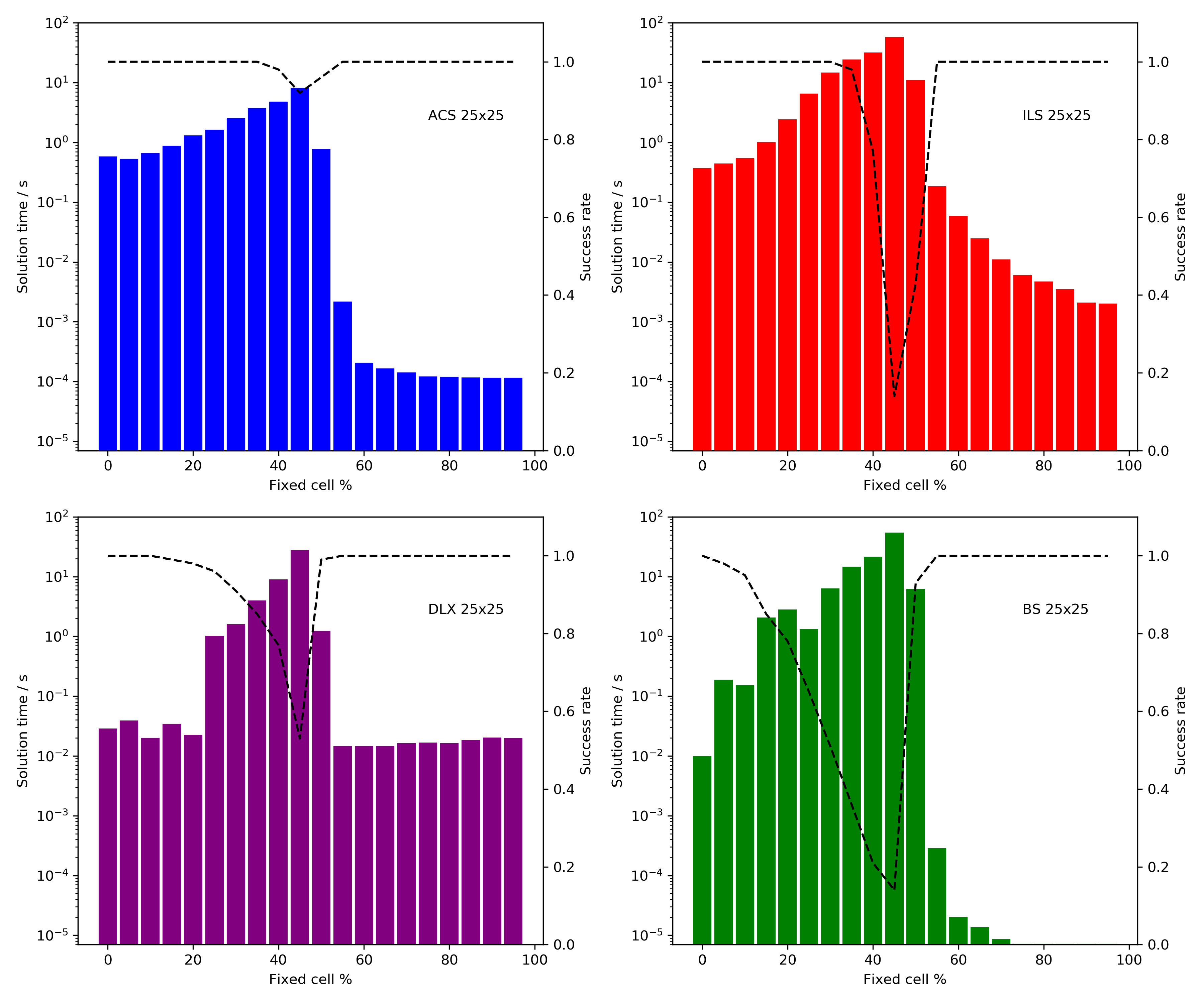}
\caption{Plots of solution time (bars) and success rate (dashed line) against fixed cell percentage for ACS, ILS, BS and DLX for the $25\times 25$ general instances.}
\label{fig:gen25x25}
\end{center}
\end{figure}

Figures~\ref{fig:gen9x9}, \ref{fig:gen16x16} and \ref{fig:gen25x25}
show the results for average execution time (for successful runs) and
success rate for the four algorithms. As in \cite{musliu2017hybrid}
and \cite{lewis2007metaheuristics}, we observe a ``phase transition" in the
difficulty of the instances as a function of the fixed cell fraction;
the difficulty is markedly greater at fixed cell fractions of around
$40-50\%$. For low values of the fixed cell fraction, the search space
is large, but there also exist many possible solutions.  As the grid
becomes denser, the size of search space decreases as well as the
number of possible solutions. At around $45\%$, the combination of
rarity of solutions and the size of the search space leads to a sharp
peak in difficulty.  

The most difficult puzzles are the $25\times 25$ instances with a
fixed cell fraction of $45\%$.  For these instances, ACS outperforms
the other three algorithms by a {\it significant} margin; ACS achieves a success
rate of $92\%$ (compared to $14\%$ for ILS, $53\%$ for DLX and $14\%$
for BS) with an average solution time of $7.9$ s (compared to $56.2$ s
for ILS, $27.1$ s for DLX and $53$ s for BS). It is interesting to
note the difference in performance between ACS and BS. These two codes
use the same underlying problem representation and constraint
propagation code; the only difference between them is the search strategy. This
comparison is compelling evidence that ACS is very efficient at
searching the solution space, giving markedly improved performance
over an exhaustive search strategy using the same underlying evaluation
routines.

\begin{figure}
\begin{center}
\includegraphics[width=\textwidth]{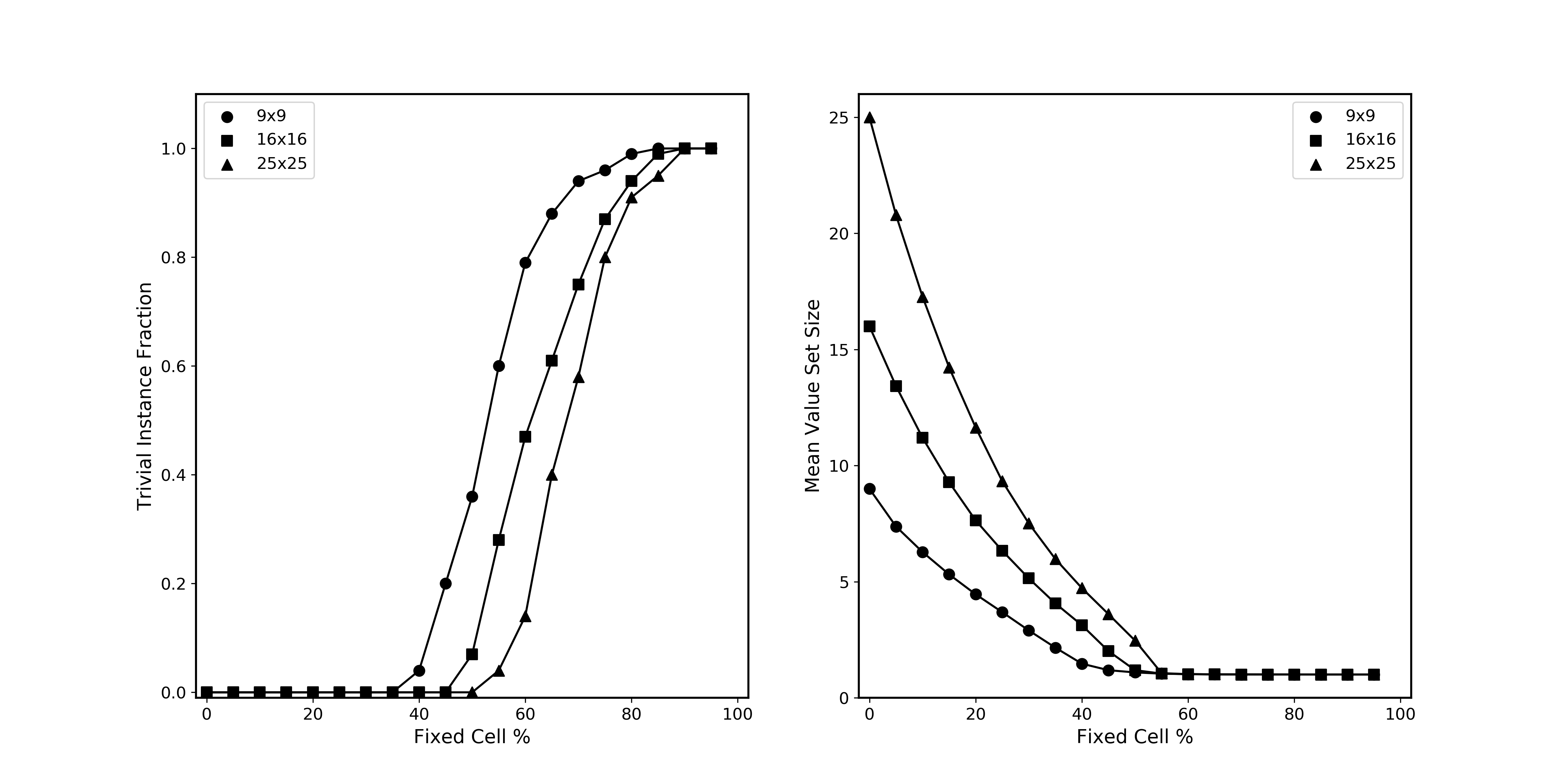}
\caption{Fraction of trivial instances and mean value set size for the test instances used for the experimental runs.}
\label{fig:instancedata}
\end{center}
\end{figure}

We analyzed the general instances to determine how many of them are ``trivial" (that is, they are solved by one application of the constraints) and the mean size of the initial value sets. Figure~\ref{fig:instancedata}
shows the fraction of instances at each fixed cell percentage which were found to be trivial, and the
mean value set size per cell after the initial application of constraints. We find that there are no trivial instances
for fixed cell percentages less than $55\%$ for the $25 \times 25$ instances, $50\%$ for $16 \times 16$, and $40\%$ for
$9\times 9$. For the hardest puzzles ($25 \times 25, 45\%$), the mean initial value set size per cell is $3.6$. 
We note that, at higher fixed cell fractions, the instances are mostly -- or entirely --
trivial, which explains the levelling off of the average solution time for ACS and BS in the timing plots.

\subsection{Evaluation of Best Value Evaporation}

In order to evaluate the effectiveness of BVE as an anti-stagnation mechanism, we ran the
same set of experiments as in Sections~\ref{sec:exp_logic} and \ref{sec:exp_gen} using the
ACS algorithm, but with best-value evaporation disabled ($f_{BVE} = 0$). For the named $9 \times 9$
logic-solvable instances, we find that ACS without BVE performs very poorly on the harder instances,
failing to solve these in most cases (see Table~\ref{tab:bve_vs_acs}). Performance on the
ten instances from \cite{sabuncu2015work} is similar to BVE, with the exception of
{\em sabuncu6}, with a success rate of $95\%$. This shows that these ten instances
are not sufficiently difficult to provide a good benchmark for solution algorithms: the search
space after applying constraints is either too small or, as is the case for four of the instances,
non-existent. 

\begin{figure}
\begin{center}
\includegraphics[width=\textwidth]{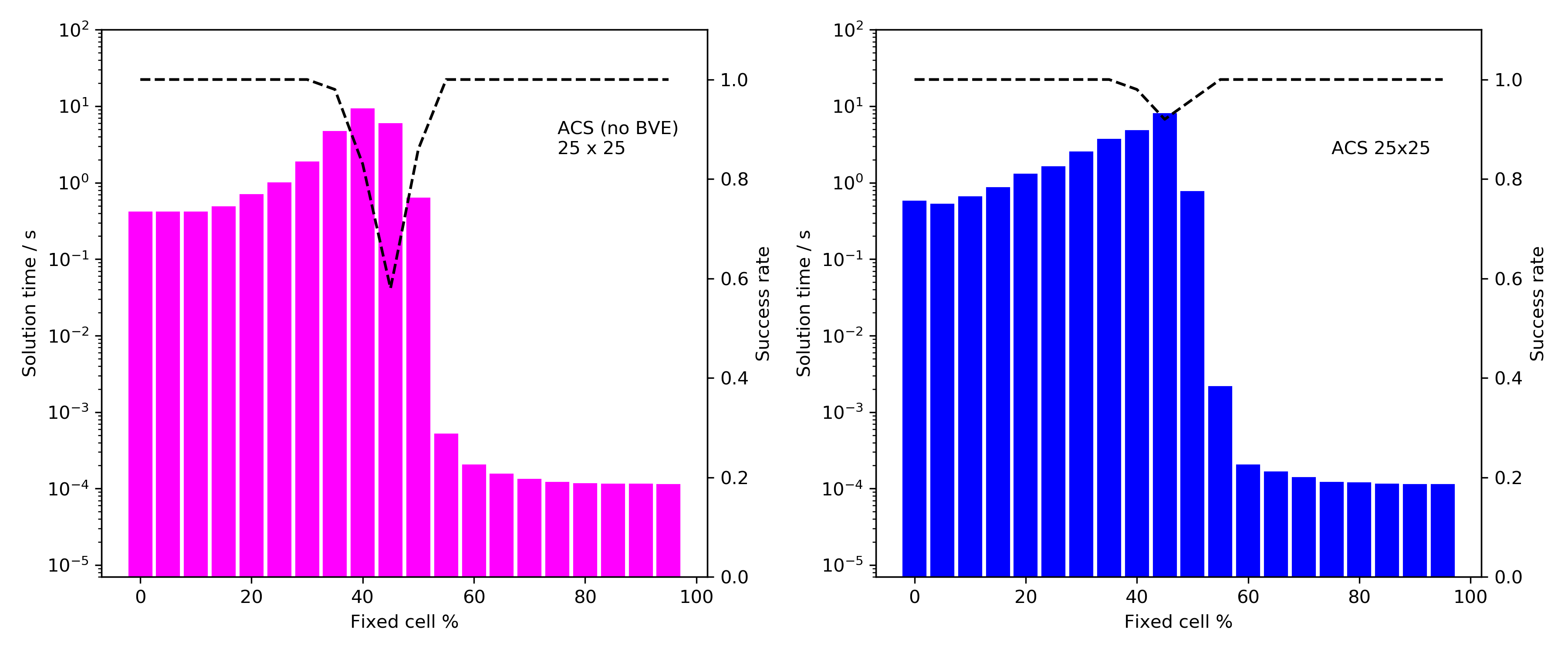}
\caption{Plot of solution time (bars) and success rate (dashed line) against fixed cell percentage for ACS without Best Value Evaporation on the $25\times 25$ general instances (left), compared to the results with BVE (right, repeated here for comparison).}
\label{fig:nobve_25}
\end{center}
\end{figure}

\begin{table}
  \begin{center}
  \caption{Performance of ACS without BVE on $9\times 9$ logic-solvable instances,
    compared to ACS with BVE. Success\% is the number of successful solutions found (with a 5 second
    timeout) in 100 runs, and $t$ is the average solution time in milliseconds.}
  \label{tab:bve_vs_acs}
  \begin{tabular}{rccccc}
    & \multicolumn{2}{c}{No BVE} & \multicolumn{2}{c}{BVE}\\
    Instance & Success\% & t/ms & Success\% & t/ms \\
    \hline
{\em sabuncu1} & 100 & 0.029  & 100 & 0.031  \\    
{\em sabuncu2} & 100 & 0.020  & 100 & 0.022 \\
{\em sabuncu3} & 100 & 0.642  & 100 & 0.627 \\
{\em sabuncu4} & 100 & 0.238  & 100 & 0.235 \\
{\em sabuncu5} & 100 & 0.017  & 100 & 0.017 \\
{\em sabuncu6} & 95  & 640.0  & 100 & 5.921 \\
{\em sabuncu7} & 100 & 4.07   & 100 & 0.456 \\
{\em sabuncu8} & 100 & 0.310  & 100 & 0.361 \\
{\em sabuncu9} & 100 & 0.634  & 100 & 0.923 \\
{\em sabuncu10} & 100 & 0.021  & 100 & 0.027 \\
{\em aiescargot} &  64 & 1223.2 & 100 & 11.11 \\
{\em platinumblond} &  14 & 892.2  & 100 & 75.75 \\
{\em goldennugget} &  36 & 866.7  & 100 & 22.95 \\
{\em reddwarf} &  34 & 1020.7 & 100 & 23.14 \\
{\em coly013} &  25 & 1320.2 & 100 & 37.95 \\
{\em tarx0134} &  42 & 1086.8 & 100 & 16.40\\
\hline
  \end{tabular}
  \end{center}
  \end{table}

Figure~\ref{fig:nobve_25} shows the results for the general $25\times 25$ instances. We see that the performance
of ACS is {\it significantly degraded} without the BVE operator. The mean solution time is similar, but the number
of failures is significantly higher; for the 45\% fixed cell instances, the success rate is $58\%$, compared to
$92\%$ with BVE enabled. The average solution time for these instances is $5.8$s, well inside the timeout of $120$s, suggesting
that the failures are due to the search stagnating at a local minimum.

\newpage
\section{Conclusions}
\label{sec:conclusions}

In this paper we presented a new algorithm for the Sudoku puzzle, based on Ant Colony Optimisation. Our method includes a new operator, which we call {\it Best Value Evaporation}, and we show that this addition to the base algorithm is essential for the prevention of premature convergence (or stagnation) of solutions. Experimental results show that our new algorithm significantly out-performs existing algorithms on large instances of Sudoku, and we provide evidence that our method provides a much more efficient search of the solution space than traditional backtracking algorithms. Future work will focus on investigating the broader applicability of our Best Value Evaporation operator to general Ant Colony Optimisation.

\bibliography{./acosudoku}

\begin{thebibliography}{10}

\bibitem{Birattari2004}
Mauro Birattari.
\newblock On the estimation of the expected performance of a metaheuristic on a
  class of instances. how many instances, how many runs?
\newblock Technical Report TR/IRIDIA/2004-001, IRIDIA, Universit{\'e} Libre de
  Bruxelles, Brussels, Belgium, 2004.

\bibitem{colbourn1984complexity}
Charles~J Colbourn.
\newblock The complexity of completing partial {L}atin squares.
\newblock {\em Discrete Applied Mathematics}, 8(1):25--30, 1984.

\bibitem{crawford2008using}
Broderick Crawford, Mary Aranda, Carlos Castro, and Eric Monfroy.
\newblock Using constraint programming to solve {S}udoku puzzles.
\newblock In {\em Third International Conference on Convergence and Hybrid
  Information Technology (ICCIT)}, volume~2, pages 926--931. IEEE, 2008.

\bibitem{delahaye2006science}
Jean-Paul Delahaye.
\newblock The science behind {S}udoku.
\newblock {\em Scientific American}, 294(6):80--87, 2006.

\bibitem{deng2013novel}
Xiu~Qin Deng and Yong Da~Li.
\newblock A novel hybrid genetic algorithm for solving {S}udoku puzzles.
\newblock {\em Optimization Letters}, 7(2):241--257, 2013.

\bibitem{dorigo2011ant}
Marco Dorigo and Mauro Birattari.
\newblock Ant colony optimization.
\newblock In {\em Encyclopedia of Machine Learning}, pages 36--39. Springer,
  2011.

\bibitem{dorigo1999ant}
Marco Dorigo and Gianni Di~Caro.
\newblock Ant colony optimization: a new meta-heuristic.
\newblock In {\em Proceedings of the 1999 Congress on Evolutionary Computation
  (CEC)}, volume~2, pages 1470--1477. IEEE, 1999.

\bibitem{dorigo1997ant}
Marco Dorigo and Luca~Maria Gambardella.
\newblock Ant colony system: a cooperative learning approach to the {T}raveling
  {S}alesman {P}roblem.
\newblock {\em IEEE Transactions on Evolutionary Computation}, 1(1):53--66,
  1997.

\bibitem{dorigo1996ant}
Marco Dorigo, Vittorio Maniezzo, and Alberto Colorni.
\newblock Ant system: optimization by a colony of cooperating agents.
\newblock {\em IEEE Transactions on Systems, Man, and Cybernetics, Part B
  (Cybernetics)}, 26(1):29--41, 1996.

\bibitem{ercsey-ravasz2012}
Maria Ercsey-Ravasz and Zoltan Toroczkai.
\newblock The chaos within {S}udoku.
\newblock {\em Sci. Rep.}, 2:725--733, 2012.

\bibitem{fletcher2007taking}
Sarah Fletcher, Frederick Johnson, and David~R Morrison.
\newblock Taking the mystery out of {S}udoku difficulty: an {O}racular model.
\newblock {\em UMAP Journal}, 29(3):327--341, 2007.

\bibitem{gareyjohnson79}
Michael~R Garey and David~S Johnson.
\newblock {\em Computers and Intractability: A Guide to the Theory of
  {NP}-Completness}.
\newblock WH Freeman: New York, 1979.

\bibitem{gunther2012entropy}
Jake Gunther and Todd Moon.
\newblock Entropy minimization for solving {S}udoku.
\newblock {\em IEEE Transactions on Signal Processing}, 60(1):508--513, 2012.

\bibitem{hereford2008integer}
James~M Hereford and Hunter Gerlach.
\newblock Integer-valued particle swarm optimization applied to {S}udoku
  puzzles.
\newblock In {\em IEEE Swarm Intelligence Symposium (SIS)}, pages 1--7. IEEE,
  2008.

\bibitem{hunt2007difficulty}
Martin Hunt, Christopher Pong, and George Tucker.
\newblock Difficulty-driven {S}udoku puzzle generation.
\newblock {\em UMAP Journal}, 29(3):343--361, 2007.

\bibitem{aiescargot}
Arto Inkala.
\newblock {\em {AI} {E}scargot - The Most Difficult Sudoku Puzzle}.
\newblock Lulu.com, Finland, 2007.

\bibitem{karimi2010sudoku}
Zahra Karimi-Dehkordi, Kamran Zamanifar, Ahmad Baraani-Dastjerdi, and Nasser
  Ghasem-Aghaee.
\newblock {S}udoku using parallel simulated annealing.
\newblock In {\em International Conference in Swarm Intelligence (ICSI)}, pages
  461--467. Springer, 2010.

\bibitem{karp1972reducibility}
Richard~M Karp.
\newblock Reducibility among combinatorial problems.
\newblock In {\em Complexity of Computer Computations}, pages 85--103.
  Springer, 1972.

\bibitem{knuth2000dancing}
Donald~E Knuth.
\newblock Dancing links.
\newblock {\em arXiv preprint cs/0011047}, 2000.

\bibitem{dlx-cpp}
Johannes Laire.
\newblock dlx-cpp.
\newblock Available at https://github.com/jlaire/dlx-cpp, accessed April 23,
  2018.

\bibitem{lewis2007metaheuristics}
Rhyd Lewis.
\newblock Metaheuristics can solve {S}udoku puzzles.
\newblock {\em Journal of Heuristics}, 13(4):387--401, 2007.

\bibitem{lopez2016ant}
Manuel L{\'o}pez-Ib{\'a}{\~n}ez, Thomas St{\"u}tzle, and Marco Dorigo.
\newblock Ant colony optimization: A component-wise overview.
\newblock {\em Handbook of Heuristics}, pages 1--37, 2016.

\bibitem{mantere2013improved}
Timo Mantere.
\newblock Improved ant colony genetic algorithm hybrid for {S}udoku solving.
\newblock In {\em Third World Congress on Information and Communication
  Technologies (WICT)}, pages 274--279. IEEE, 2013.

\bibitem{mantere2007solving}
Timo Mantere and Janne Koljonen.
\newblock Solving, rating and generating {S}udoku puzzles with {GA}.
\newblock In {\em IEEE Congress on Evolutionary Computation (CEC)}, pages
  1382--1389. IEEE, 2007.

\bibitem{moraglio2007geometric}
Alberto Moraglio and Julian Togelius.
\newblock Geometric particle swarm optimization for the {S}udoku puzzle.
\newblock In {\em Proceedings of the 9th Annual Conference on Genetic and
  Evolutionary Computation (GECCO)}, pages 118--125. ACM, 2007.

\bibitem{musliu2017hybrid}
Nysret Musliu and Felix Winter.
\newblock A hybrid approach for the {S}udoku problem: using constraint
  programming in iterated local search.
\newblock {\em IEEE Intelligent Systems}, 32(2):52--62, 2017.

\bibitem{norvigsolving}
Peter Norvig.
\newblock Solving every {S}udoku puzzle.
\newblock Available at http://norvig.com/sudoku.html, accessed March 13, 2018.

\bibitem{pacurib2009solving}
Jaysonne~A Pacurib, Glaiza Mae~M Seno, and John Paul~T Yusiong.
\newblock Solving {S}udoku puzzles using improved artificial bee colony
  algorithm.
\newblock In {\em Fourth International Conference on Innovative Computing,
  Information and Control (ICICIC)}, pages 885--888. IEEE, 2009.

\bibitem{sabuncu2015work}
Ibrahim Sabuncu.
\newblock Work-in-progress: solving {S}udoku puzzles using hybrid ant colony
  optimization algorithm.
\newblock In {\em 1st International Conference on Industrial Networks and
  Intelligent Systems (INISCom)}, pages 181--184. IEEE, 2015.

\bibitem{schiff2015ant}
Krzysztof Schiff.
\newblock An ant algorithm for the {S}udoku problem.
\newblock {\em Journal of Automation, Mobile Robotics and Intelligent Systems},
  9, 2015.

\bibitem{segura2016importance}
Carlos Segura, S~Ivvan~Valdez Pe{\~n}a, Salvador~Botello Rionda, and
  Arturo~Hern{\'a}ndez Aguirre.
\newblock The importance of diversity in the application of evolutionary
  algorithms to the {S}udoku problem.
\newblock In {\em IEEE Congress on Evolutionary Computation (CEC)}, pages
  919--926. IEEE, 2016.

\bibitem{soto2013hybrid}
Ricardo Soto, Broderick Crawford, Cristian Galleguillos, Eric Monfroy, and
  Fernando Paredes.
\newblock A hybrid ac3-tabu search algorithm for solving {S}udoku puzzles.
\newblock {\em Expert Systems with Applications}, 40(15):5817--5821, 2013.

\bibitem{wang2015evolutionary}
Zhiwen Wang, Toshiyuki Yasuda, and Kazuhiro Ohkura.
\newblock An evolutionary approach to {S}udoku puzzles with filtered mutations.
\newblock In {\em IEEE Congress on Evolutionary Computation (CEC)}, pages
  1732--1737. IEEE, 2015.

\bibitem{weber2005sat}
Tjark Weber.
\newblock A {SAT}-based {S}udoku solver.
\newblock In Geoff Sutcliffe and Andrei Voronkov, editors, {\em The 12th
  International Conference on Logic for Programming, Artificial Intelligence,
  and Reasoning (LPAR): Short Paper Proceedings}, pages 11--15, 2005.

\bibitem{yato2003complexity}
Takayuki Yato and Takahiro Seta.
\newblock Complexity and completeness of finding another solution and its
  application to puzzles.
\newblock {\em IEICE Transactions on Fundamentals of Electronics,
  Communications and Computer Sciences}, 86(5):1052--1060, 2003.

\end{thebibliography}

\end{document}